%% file: samplepaper.tex
\documentclass[runningheads]{llncs}
\usepackage{comment}
\usepackage{amsmath,amsfonts,amssymb}
\usepackage{algorithmic}
\usepackage{algorithm}
\usepackage{array}
\usepackage{textcomp}
\usepackage{stfloats}
\usepackage{url}
\usepackage{verbatim}
\usepackage{graphicx}
\usepackage{caption}
\usepackage{subcaption}
\usepackage{cite}
\usepackage{hyperref}
\usepackage[nameinlink,noabbrev]{cleveref}
\usepackage{booktabs}
\usepackage{xcolor}
\usepackage{etoolbox}

\definecolor{cvprblue}{rgb}{0.21,0.49,0.74}

\def\method{CreatiPoster}

\makeatletter
\newcommand{\authorfootnotetext}[2]{%
  \insert\footins{%
    \reset@font\footnotesize
    \interlinepenalty\interfootnotelinepenalty
    \splittopskip\footnotesep
    \splitmaxdepth\dp\strutbox \floatingpenalty\@MM
    \hsize\columnwidth\@parboxrestore
    \color@begingroup
      \parindent\fnindent\leftskip\fnindent\noindent
      \llap{\hb@xt@1em{\hss$#1$\ }}\ignorespaces#2\@finalstrut\strutbox
    \color@endgroup}%
}
\makeatother

\makeatletter
\renewcommand\section{\@startsection{section}{1}{\z@}%
                       {-11\p@ \@plus -3\p@ \@minus -3\p@}%
                       {5\p@ \@plus 2\p@ \@minus 2\p@}%
                       {\normalfont\large\bfseries\boldmath
                        \rightskip=\z@ \@plus 8em\pretolerance=10000 }}
\renewcommand\subsection{\@startsection{subsection}{2}{\z@}%
                       {-10\p@ \@plus -3\p@ \@minus -3\p@}%
                       {3\p@ \@plus 2\p@ \@minus 2\p@}%
                       {\normalfont\normalsize\bfseries\boldmath
                        \rightskip=\z@ \@plus 8em\pretolerance=10000 }}
\renewcommand\paragraph{\@startsection{paragraph}{4}{\z@}%
                       {-6\p@ \@plus -3\p@ \@minus -3\p@}%
                       {-0.5em \@plus -0.22em \@minus -0.1em}%
                       {\normalfont\normalsize\itshape}}
\makeatother

\begin{document}
\title{CreatiPoster: Towards Editable and Controllable Multi-Layer Graphic Design Generation}
\titlerunning{CreatiPoster: Editable Multi-Layer Graphic Design Generation}
\author{Dexiang Hong\inst{1}\unskip$^\star$ \and
Zhao Zhang\inst{2}\unskip$^\star$ \and
Weidong Chen\inst{1}\unskip$^\dagger$ \and
Yutao Cheng\inst{2} \and
Maoke Yang\inst{2} \and
Gonglei Shi\inst{2} \and
Hui Zhang\inst{3} \and
Zhendong Mao\inst{1}}
\authorrunning{D. Hong et al.}
%
\institute{University of Science and Technology of China, Hefei, China \and
Intelligent Creation \and
Fudan University, Shanghai, China\\
}
\maketitle              
\authorfootnotetext{\star}{Equal contribution.}
\authorfootnotetext{\dagger}{Corresponding author.}
\vspace{-0.35cm}

\begin{abstract}
Graphic design is central to communication, yet producing high-quality and editable compositions remains time-consuming. Existing AI tools struggle to faithfully place user assets, preserve editability, or reach professional quality, while commercial systems depend on vast, hard-to-replicate template libraries.
We introduce \method{}, an open framework that reframes graphic design generation as \emph{editable design program synthesis}.
\method{} follows a two-stage pipeline. A \emph{layered protocol generator}---an RGBA multimodal model---first produces a JSON program that specifies every text and asset layer (content, geometry, style, and order) plus a background caption, which is rendered into an editable foreground; a \emph{foreground-aware background synthesizer}, built on an MM-DiT backbone, then generates only the background. Conditioned on the rendered foreground, it produces a background that harmonizes with the foreground in color, layout, and style.
Because a design is represented as an editable program of native layers rather than baked-in pixels, user assets stay intact and every element remains editable after generation, while layout learned from data removes any dependence on template libraries.
On a new benchmark with automated metrics, \method{} surpasses leading open-source and proprietary commercial systems.
It further supports diverse applications---canvas editing, text overlay, responsive resizing, multilingual adaptation, and animated posters---advancing the democratization of AI-assisted graphic design.
\end{abstract}

\begin{figure}[tbp]
\centering
\includegraphics[width=0.85\textwidth]{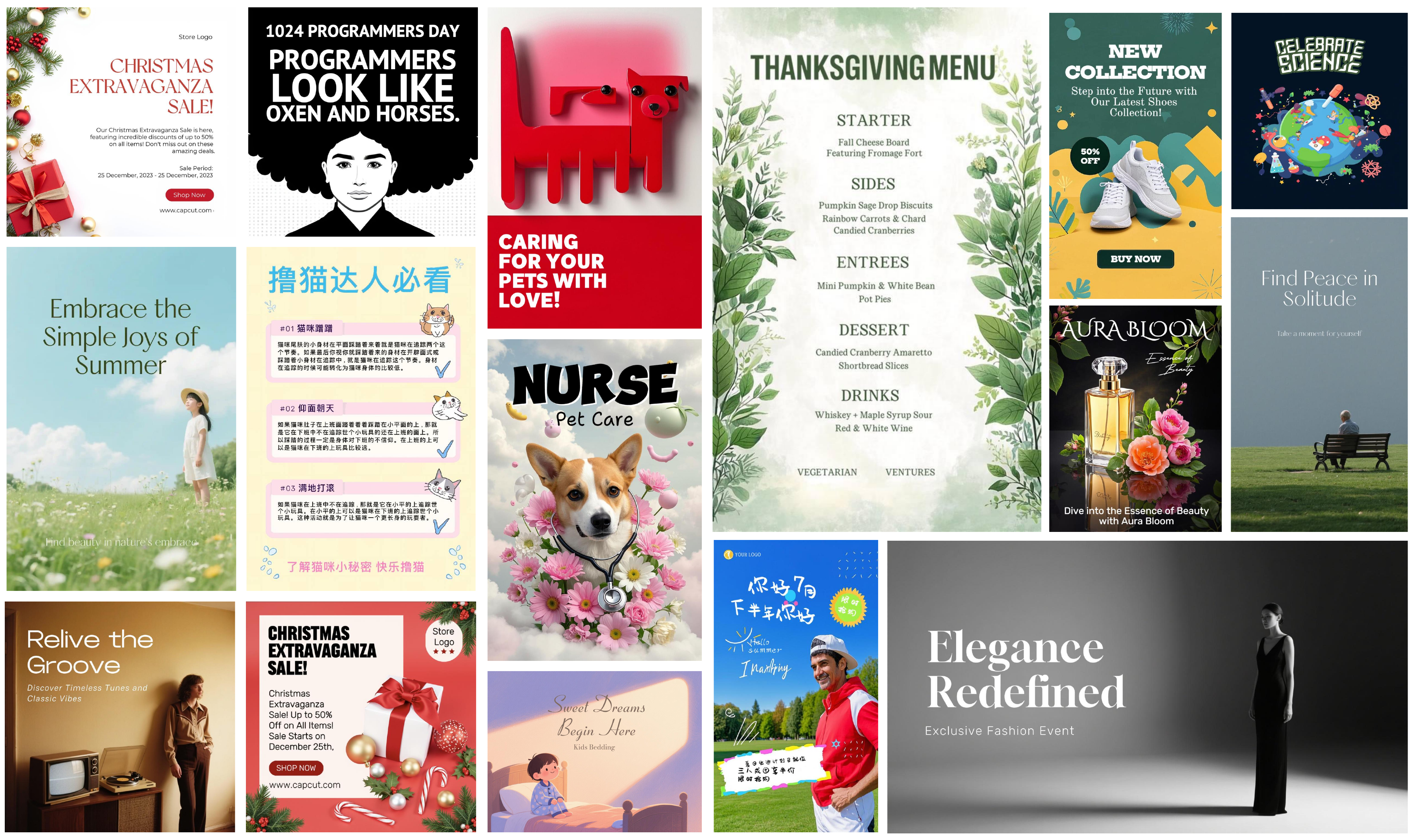}
\caption{Multi-layer compositions produced by \method. The JSON design program lists every text and asset layer, letting users freely edit content, placement, and style in the GUI editor.}
\label{fig:teaser}
\end{figure}

\input{sections/introduction}

\input{sections/related_work}
\input{sections/method}

\input{sections/experiments}
\input{sections/application}
\input{sections/conclusion}

{\sloppy\noindent\textbf{Acknowledgement.} This research is supported by Artificial Intelligence-National Science and Technology Major Project 2023ZD0121200, and the National Natural Science Foundation of China under Grants 62302474.\par}

%
%
%
\bibliographystyle{splncs04}
\bibliography{main}
\end{document}

%% file: sections/introduction.tex
\section{Introduction}
\label{sec:introduction}
Graphic design combines text and images to deliver targeted messages in posters, social media graphics, and business cards. It demands expertise and tool mastery, and aesthetically refined compositions are especially time-consuming for non-experts.
AI has recently been integrated across stages of the design process---layout generation~\cite{yamaguchi2021crello}, hierarchical layout~\cite{cheng2025graphist}, logo creation~\cite{wang2022aesthetic}, artistic text~\cite{gao2019artistic}, and color harmony~\cite{cohen2006color}---and some recent methods~\cite{jia2023cole,inoue2024opencole} generate complete designs from language instructions. These approaches, however, still struggle to incorporate user-provided assets and to reach professional aesthetic standards. Commercial platforms such as Canva Magic Design, Adobe Express, and Microsoft Designer\footnote{\url{https://www.canva.com/magic-design/}, \url{https://www.adobe.com/express/}, \url{https://designer.microsoft.com/}} pair vast template libraries with intelligent search and auto-copywriting, but building such libraries is impractical for most individuals and small organizations.

A high-quality AI-generated graphic composition should jointly satisfy four criteria: \underline{text accuracy} (correct spelling and typography), \underline{asset fidelity} (user-supplied images, photos, or logos must be preserved and well-placed), \underline{editability} (users must be able to modify text and replace assets), and \underline{aesthetic appeal} (the design should be visually pleasing). No existing AI system meets all four simultaneously.

To fill this research gap, we develop an open framework, \method, that recasts graphic design generation as \emph{editable design program synthesis} rather than flattened image synthesis.
As shown in \Cref{fig:pipeline}, \method{} couples a \emph{layered protocol generator}---a large multimodal network~\cite{cheng2025graphist}---with a \emph{foreground-aware background synthesizer}. The protocol generator turns user instructions, optional RGBA assets, a target canvas size, and optional user constraints into an executable JSON design program that lists every text or asset layer with its semantic, geometric, stylistic, and ordering attributes, together with a concise background caption. This program can be rendered immediately with engines such as Skia\footnote{\url{https://skia.org/}} into fully editable foreground layers (see supplementary). The background synthesizer then generates only the background layer conditioned on the rendered foreground and the background caption, producing a complete multi-layer composition whose elements remain disentangled for later editing.
Crucially, the protocol generator is trained under a single \emph{protocol completion} objective: prompt-only, asset-conditioned, canvas-conditioned, and re-layout generation are all instantiated as completing a partially observed protocol, without task-specific heads. Evaluated on a new benchmark with automated metrics, \method{} outperforms existing open-source and commercial systems, and we further release $100{,}000$ copyright-free multi-layer samples to catalyze research in AI-driven graphic design.
The contributions of this paper are summarized as follows:

\begin{itemize}
    \item \textbf{Editable design program synthesis.}
    We formulate graphic design generation as editable design program synthesis and propose CreatiPoster, an open framework that produces editable multi-layer JSON programs instead of flattened images.

    \item \textbf{Layered protocol generation.}
    We develop an RGBA multimodal protocol generator that specifies the content, geometry, style, and order of each text or asset layer, together with a background caption.

    \item \textbf{Foreground-aware background synthesis.}
    We introduce an MM-DiT-based background synthesizer that generates only the background conditioned on the rendered foreground, preserving text accuracy, asset fidelity, and editability.

    \item \textbf{Evaluation and applications.}
    Experiments on a new benchmark show that CreatiPoster outperforms leading open-source and commercial systems, while its editable representation enables canvas editing, text overlay, responsive resizing, multilingual adaptation, and animated posters.
\end{itemize}

%% file: sections/related_work.tex
\section{Related Work}
\label{sec:related_work}
\paragraph{Large Multimodal Models.}
Large multimodal models (LMMs)~\cite{li2023multimodal,zhang2024mm} have been widely adopted for video understanding~\cite{tang2023video}, OS agents~\cite{wang2024mobileagent}, image generation~\cite{zhou2024transfusion}, embodied intelligence~\cite{driess2023palme}, audio-visual understanding~\cite{qin2025queryBasedCollaborative,meng2026audioVisualExchangeAware}, and affective reasoning~\cite{song2026bridgingSubjectivity,guo2025emoverse}. In \method, we use an LMM to predict the layout and attributes of text and user assets. LMMs have also made notable progress in coordinate representation~\cite{peng2023kosmos2}, with Shikra~\cite{chen2023shikra} expressing spatial positions as numeric tokens in natural language, and Graphist~\cite{cheng2025graphist} extending this idea to layout generation.

\paragraph{Image Generation.}
Text-to-image generation has advanced rapidly. Recent MM-DiT models---Stable Diffusion 3.x~\cite{esser2024sd3}, FLUX.1~\cite{flux}, and Seedream~\cite{gao2025seedream3.0}---scale parameters and data, while conditional control~\cite{tan2024ominicontrol,zhang2024creatilayout} adds subject and layout grounding. LayerDiffuse~\cite{zhang2024transparent} first enabled multi-layer conditional generation.

\paragraph{Design Systems.}
Automated graphic design should look good, communicate clearly, support user-defined assets, and remain editable. Early systems relied on aesthetic rules~\cite{bauerly2006computational_aesthetics} or constraint-based optimization~\cite{hurst2009review_layout}, and later work tackled sub-problems such as layout~\cite{li2019layoutgan,inoue2023FlexDM,cheng2025graphist,jiang2023layoutformer++} and color harmony~\cite{cohen2006color}. More recent approaches simplify the modeling, but typically sacrifice either editability or asset support: layout-based pipelines~\cite{hsu2023posterlayout,seol2024posterllama} require the main image to fill the canvas; multi-stage layered methods~\cite{jia2023cole,inoue2024opencole,chen2025posta,wang2025banneragency} drop user assets; ART~\cite{pu2025art} compromises text editability; and purely generative methods~\cite{gao2025postermaker} lack editability. \method{} preserves both user-defined assets and full text/layer editability. Complementary to generation, CreatiParser~\cite{chen2026creatiparser} tackles the inverse problem of parsing a flattened raster design back into editable layers.

%% file: sections/method.tex
\section{Proposed Method}
\label{sec:method}

\subsection{Editable Design Program Formulation}
\label{sec:method_formulation}
Given a prompt $p\!\in\!\mathcal{S}$, canvas size $\mathbf{s}\!\in\!\mathbb{R}^{2}$, optional RGBA assets $\mathcal{I}=\{I_i\!\in\!\mathbb{R}^{h_i\times w_i\times 4}\}_{i=1}^{n}$, and optional constraints $\mathcal{C}$ (e.g., locked layers or specified attributes), \method{} synthesizes an \emph{editable design program} rather than a flattened image.
The framework (\Cref{fig:pipeline}) consists of a \emph{Layered Protocol Generator} $\operatorname{PM}$ and a \emph{Foreground-Aware Background Synthesizer} $\operatorname{BM}$, coupled by a rendering engine $\operatorname{R}$:
\begin{align}
\operatorname{PM}(\mathcal{I},\,p,\,\mathbf{s},\,\mathcal{C}) &\;\longrightarrow\;
\mathcal{P}= \bigl\{\mathcal{L},\; c_{\mathrm{bg}}\bigr\},
\label{eq:pm} \\
\operatorname{R}(\mathbf{s},\,\mathcal{L})     &\;\longrightarrow\;
I_{\mathrm{fg}},                                \label{eq:render} \\
\operatorname{BM}(I_{\mathrm{fg}},\,c_{\mathrm{bg}})          &\;\longrightarrow\;
I_{\mathrm{bg}},                                \label{eq:bm} \\
O &= \bigl\{\texttt{BG}\!:\!I_{\mathrm{bg}},\;\texttt{FG}\!:\!\mathcal{L}\bigr\},
\label{eq:output}
\end{align}
where $\mathcal{L}=[\ell_1,\dots,\ell_m]$ is an ordered JSON list of editable text/asset layers, $c_{\mathrm{bg}}$ is a concise background caption, and $I_{\mathrm{fg}}$, $I_{\mathrm{bg}}$ are the rendered foreground and synthesized background. This separation preserves text accuracy, asset fidelity, and editability.

\input{figs/pipeline}
\input{sections/protocol_model}
\input{sections/background_model}
\input{sections/generation_modes}

%% file: figs/pipeline.tex
\begin{figure}[tbp]
    \centering
    \makebox[\linewidth][c]{\includegraphics[width=\linewidth]{figs/pipeline.png}}
  \caption{Overview of editable design program synthesis in \method. Multimodal user inputs (prompt, optional RGBA assets, canvas size, and optional user constraints) are first turned into an executable JSON design program by the layered protocol generator. The program is rendered into editable foreground layers, on which the foreground-aware background synthesizer composes a coherent background, yielding an editable multi-layer composition rather than a flattened image.}
  \label{fig:pipeline}
\end{figure}

%% file: sections/protocol_model.tex
\subsection{Layered Protocol Generation}
\label{sec:method_protocol}
The layered protocol generator $\operatorname{PM}$ produces $\mathcal{P}$ in a single multimodal forward pass.
Following CreatiGraphist~\cite{cheng2025graphist}, it combines an RGBA encoder, visual shrinker, and LLM: the encoder preserves asset silhouettes and transparent boundaries, while the shrinker maps each asset to 64 tokens, retaining edge/global cues and keeping multiple assets, the prompt, canvas size, and constraints within one context window.

\noindent\textbf{Layered design protocol.}
The protocol $\mathcal{L}=[\ell_1,\dots,\ell_m]$ is an executable JSON specification whose layer $\ell_j$ contains four attribute groups,
\begin{equation}
\ell_j=\bigl(\,\mathbf{a}^{\mathrm{sem}}_j,\;\mathbf{a}^{\mathrm{geo}}_j,\;\mathbf{a}^{\mathrm{sty}}_j,\;o_j\,\bigr),
\label{eq:layer}
\end{equation}
where $\mathbf{a}^{\mathrm{sem}}_j$ records layer type, text, or asset identity; $\mathbf{a}^{\mathrm{geo}}_j$ records position, size, crop, rotation, bend, and masks; $\mathbf{a}^{\mathrm{sty}}_j$ records font, color, stroke, opacity, alignment, spacing, and text styles; and $o_j\!\in\!\{1,\dots,m\}$ defines the bottom-to-top order.
Unlike box-only layouts, this protocol captures placement, appearance, and inter-layer order while remaining fully editable in a GUI editor (see supplementary).

\noindent\textbf{Protocol completion training.}
We train $\operatorname{PM}$ as a \emph{protocol completion} model rather than a from-scratch generator.
Let $\widetilde{\mathcal{P}}\!=\!\mathrm{Mask}(\mathcal{P})$ denote a partial protocol formed by exposing layers $\mathcal{L}'\!\subseteq\!\mathcal{L}$ and randomly masking content, geometry, or style fields:
\begin{equation}
\mathrm{Mask}(\mathcal{P})=\bigl\{\,m_j\!\odot\!\ell_j : \ell_j\!\in\!\mathcal{L}'\,\bigr\},
\qquad
m_j\!\sim\!\mathrm{Bernoulli}(\rho),
\label{eq:mask}
\end{equation}
where $m_j$ retains each field with keep-probability $\rho$; $\rho\!=\!0$ recovers prompt-only generation and $\rho\!=\!1$ a fully specified canvas.
The model is trained to recover the full protocol $\mathcal{P}$ from the visible constraints, following the multi-stage training schedule of~\cite{cheng2025graphist}:
\begin{equation}
\operatorname{PM}\bigl(\mathcal{I},\,p,\,\mathbf{s},\,\widetilde{\mathcal{P}}\bigr)\;\longrightarrow\;\mathcal{P}.
\label{eq:protocol_completion}
\end{equation}
With $\mathbf{y}=\mathrm{Serialize}(\mathcal{P})=(y_1,\dots,y_T)$ and $\mathcal{X}=(\mathcal{I},p,\mathbf{s},\widetilde{\mathcal{P}})$, $\operatorname{PM}$ is optimized by autoregressive cross-entropy:
\begin{equation}
\mathcal{L}_{\mathrm{PM}}
=-\,\mathbb{E}_{(\mathcal{P},\mathcal{X})}
\sum_{t=1}^{T}
\log p_{\theta}\!\bigl(y_t \,\big|\, y_{<t},\,\mathcal{X}\bigr),
\label{eq:pm_loss}
\end{equation}
This objective covers prompt-only, asset, and canvas-editing cases without task-specific heads.
At inference, the same network supports the controllable modes in \Cref{sec:method_modes}.

%% file: sections/background_model.tex
\subsection{Foreground-Aware Background Synthesis}
\label{sec:method_background}
After rendering the editable foreground $I_{\mathrm{fg}}$, $\operatorname{BM}$ synthesizes only the background $I_{\mathrm{bg}}$ conditioned on $I_{\mathrm{fg}}$ and caption $c_{\mathrm{bg}}$, avoiding corruption of text or user assets.

\noindent\textbf{Multimodal conditioning.}
We convert the RGBA foreground to RGB on a neutral gray canvas and encode it with the native VAE into foreground tokens $h_f$; the text encoder produces caption tokens $h_b$, and noisy image tokens $h_z$ are sampled in latent space.
MM-DiT blocks concatenate the three streams and jointly attend over them. For modality $*\!\in\!\{b,z,f\}$, $Q^{*},K^{*},V^{*}\!=\!\mathrm{Linear}(h^{*})$, and:
\begin{equation}
\begin{split}
{h^{b}}',\,{h^{z}}',\,{h^{f}}' &=
\mathrm{Att}\!\Big(
[\mathbf{Q}^{b},\mathbf{Q}^{z},\mathbf{Q}^{f}],\,
[\mathbf{K}^{b},\mathbf{K}^{z},\mathbf{K}^{f}], \\
&\qquad\qquad
[\mathbf{V}^{b},\mathbf{V}^{z},\mathbf{V}^{f}]
\Big),
\end{split}
\label{eq:multimodal-attention}
\end{equation}
with scaled dot-product attention over $\mathbf{Q}\!=\![\mathbf{Q}^{b},\mathbf{Q}^{z},\mathbf{Q}^{f}]$ and likewise for $\mathbf{K},\mathbf{V}$:
\begin{equation}
\mathrm{Att}(\mathbf{Q},\mathbf{K},\mathbf{V})
=\mathrm{softmax}\!\Bigl(\tfrac{\mathbf{Q}\mathbf{K}^{\!\top}}{\sqrt{d}}\Bigr)\,\mathbf{V},
\label{eq:sdpa}
\end{equation}
where $d$ is the per-head channel dimension, allowing noisy background tokens to attend to the caption and foreground elements in one pass.
For lightweight foreground adaptation, LoRA modules and AdaLN are attached to foreground projection paths. Each adapted projection reparameterizes frozen $W_0\!\in\!\mathbb{R}^{d\times k}$ as
\begin{equation}
W=W_0+\Delta W=W_0+\tfrac{\alpha}{r}\,BA,
\qquad B\!\in\!\mathbb{R}^{d\times r},\; A\!\in\!\mathbb{R}^{r\times k},
\label{eq:lora}
\end{equation}
where only $A,B$ and $\alpha$ are trainable, leaving the backbone intact.
We also reuse the noisy-image positional encoding on foreground tokens so each foreground element informs its surrounding background region.

\noindent\textbf{Two-stage training.}
With the diffusion backbone frozen, only LoRA components are trained.
Following flow matching, a clean background latent $x_0$ and Gaussian noise $\epsilon\!\sim\!\mathcal{N}(0,\mathbf{I})$ are interpolated as $x_\tau\!=\!(1-\tau)\,x_0+\tau\,\epsilon$, and $v_\theta$ regresses velocity $\epsilon-x_0$ conditioned on foreground and caption tokens:
\begin{equation}
\mathcal{L}_{\mathrm{BM}}
=\mathbb{E}_{x_0,\epsilon,\tau}
\Bigl[\,
\bigl\|\,v_\theta\!\bigl(x_\tau,\tau,\,h_f,\,h_b\bigr)-(\epsilon-x_0)\,\bigr\|_2^{2}
\Bigr],
\label{eq:flow_matching}
\end{equation}
We pre-train at $512$ resolution with a lognormal noise schedule (mean $0.5$, standard deviation $1$), then post-train at $1024$ with a uniform schedule.
This combines low-resolution foreground sensitivity with high-resolution adaptability across poster sizes.

%% file: sections/generation_modes.tex
\subsection{Unified Controllable Generation Modes}
\label{sec:method_modes}
Because protocol completion recovers $\mathcal{P}$ from a partially observed $\widetilde{\mathcal{P}}$ plus optional prompts and assets, one $\operatorname{PM}$ supports the modes in \Cref{tab:modes} without task-specific heads.
At inference, the visible fields select the task: an empty protocol gives prompt-only generation; asset placeholders give asset-conditioned generation; partial or locked layers give canvas-conditioned editing; and a previous protocol plus a new canvas size gives re-layout.
The applications in \Cref{sec:app} therefore reuse the same formulation instead of bespoke pipelines.

\begin{table}[tbp]
\caption{Unified controllable generation modes supported by a single protocol-completion model.}
\label{tab:modes}
\centering\small
\renewcommand{\arraystretch}{1.25}
\setlength{\tabcolsep}{6pt}
\begin{tabular}{@{}
  >{\raggedright\arraybackslash}p{0.20\columnwidth}
  >{\raggedright\arraybackslash}p{0.30\columnwidth}
  >{\raggedright\arraybackslash}p{0.40\columnwidth}@{}}
\toprule
\textbf{Mode} & \textbf{Input} & \textbf{Model behavior} \\
\midrule
Prompt-only & Prompt + canvas size & Generate complete protocol and background caption \\
\addlinespace[2pt]
Asset-conditioned & Prompt + RGBA assets & Preserve uploaded assets and predict placement, crop, scale, and style \\
\addlinespace[2pt]
Canvas-conditioned & Prompt + partial protocol / locked layers & Complete missing layers or attributes under the visible constraints \\
\addlinespace[2pt]
Re-layout & Previous protocol + new canvas size & Regenerate layout while preserving content and style \\
\bottomrule
\end{tabular}
\end{table}

%% file: sections/experiments.tex
\section{Experiments}
\label{sec:experiments}

\subsection{Implementation Details}
We use InternLM2.5~\cite{cai2024internlm2} as the LLM backbone of the layered protocol generator, training it on a mixture of in-house designer poster data, multimodal content understanding data, and conversational data.
For the background synthesizer we train two variants: \method-F (FLUX-dev~\cite{flux} backbone) and \method-S (in-house backbone); both use a LoRA rank of $256$.
Training runs on $8$ NVIDIA A100 GPUs, taking about five days for the protocol generator and three days for the background synthesizer.

\subsection{Benchmark and Metrics}
Our test set contains $45$ prompt-only cases (poster images collected online with captions generated by a multimodal LLM\footnote{\url{https://www.doubao.com/chat/}}), $39$ prompt-with-single-asset cases, and $6$ prompt-with-multi-asset cases (assets extracted from text-to-image generations and re-captioned by the same LLM).
Standard metrics for natural image synthesis such as FID, PSNR, and SSIM are sensitive to the underlying generator and do not capture graphic design quality. Following advice from professional designers, we evaluate four dimensions---\emph{Layout}, \emph{Color}, \emph{Graphic Style}, and \emph{Compliance}---summarized in \Cref{tab:evaluation_dimension}. We score each case with GPT-4.1\footnote{\url{https://openai.com/index/gpt-4-1/}} on a $1$--$5$ scale, sampling ten times per case and taking the majority vote, and additionally collect a single-blind $1$--$5$ overall rating from $10$ human volunteers.
We compare against the open-source OpenCOLE~\cite{inoue2024opencole} and the closed-source commercial systems Microsoft Designer and Canva Magic Design.

\begin{table}[tbp]
\caption{Graphic design evaluation dimensions.}
\label{tab:evaluation_dimension}
\centering\small
\begin{tabular}{p{0.18\columnwidth}p{0.74\columnwidth}}
    \toprule
    \textbf{Dimension} & \textbf{Description} \\
    \midrule
    Layout & Compositional appropriateness and arrangement of elements. \\
    Color & Whether the palette suits the content and is internally harmonious. \\
    Graphic Style & Coherence among fonts, decorative elements, assets, and background. \\
    Compliance & Faithfulness to the user prompt. \\
    \bottomrule
\end{tabular}
\end{table}

\subsection{Results and Discussion}
\begin{figure}[!t]
\centering
\captionsetup[subfigure]{skip=1pt,font=small}
\begin{subcaptionbox}{Prompt-only setting.\label{fig:comp_woimg}}[\linewidth]
{\includegraphics[width=\linewidth,trim=0 24bp 0 0,clip]{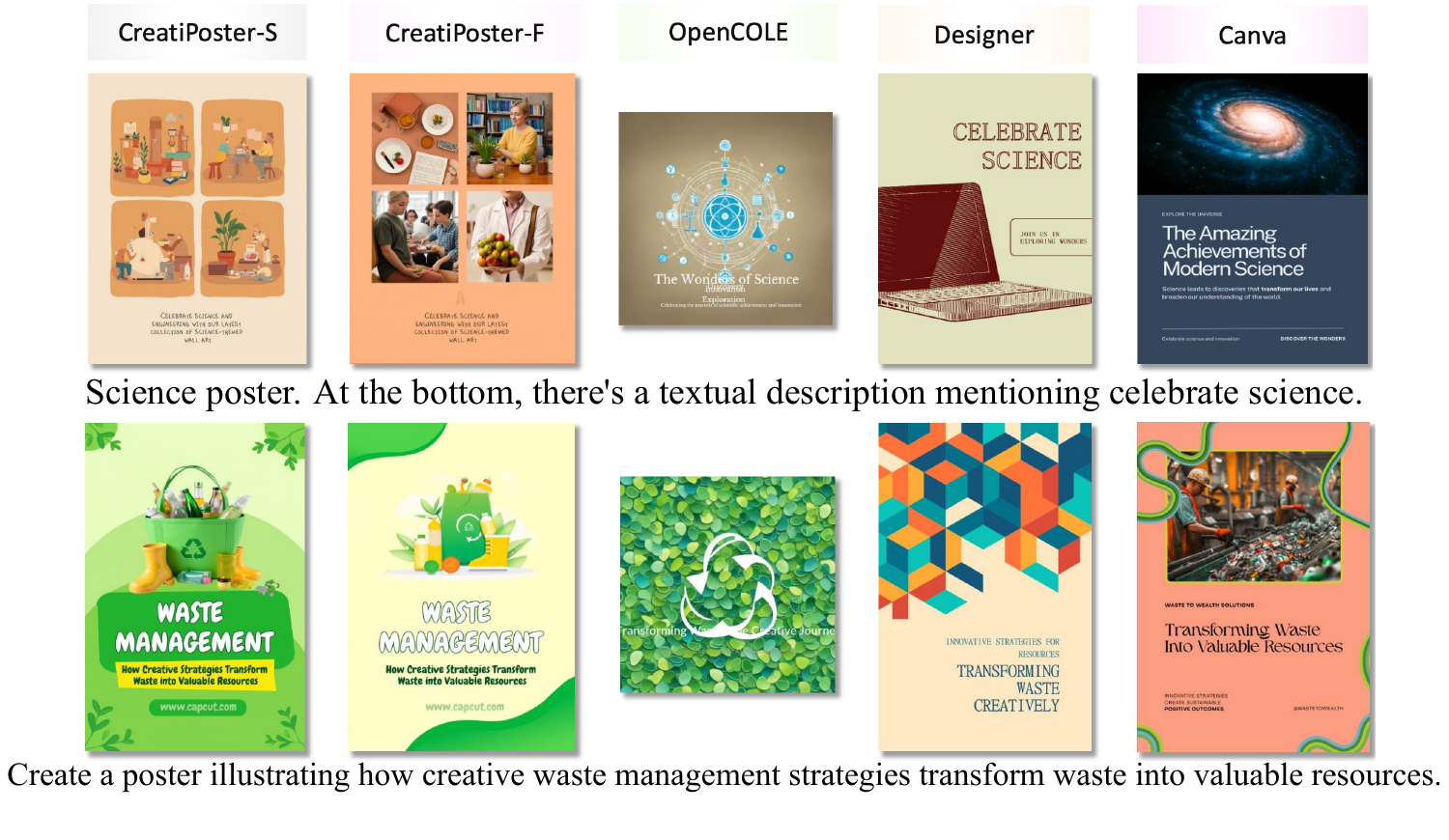}}
\end{subcaptionbox}\\[1pt]
\begin{subcaptionbox}{Prompt-with-asset setting.\label{fig:comp_wimg}}[\linewidth]
{\includegraphics[width=\linewidth,trim=0 31bp 0 0,clip]{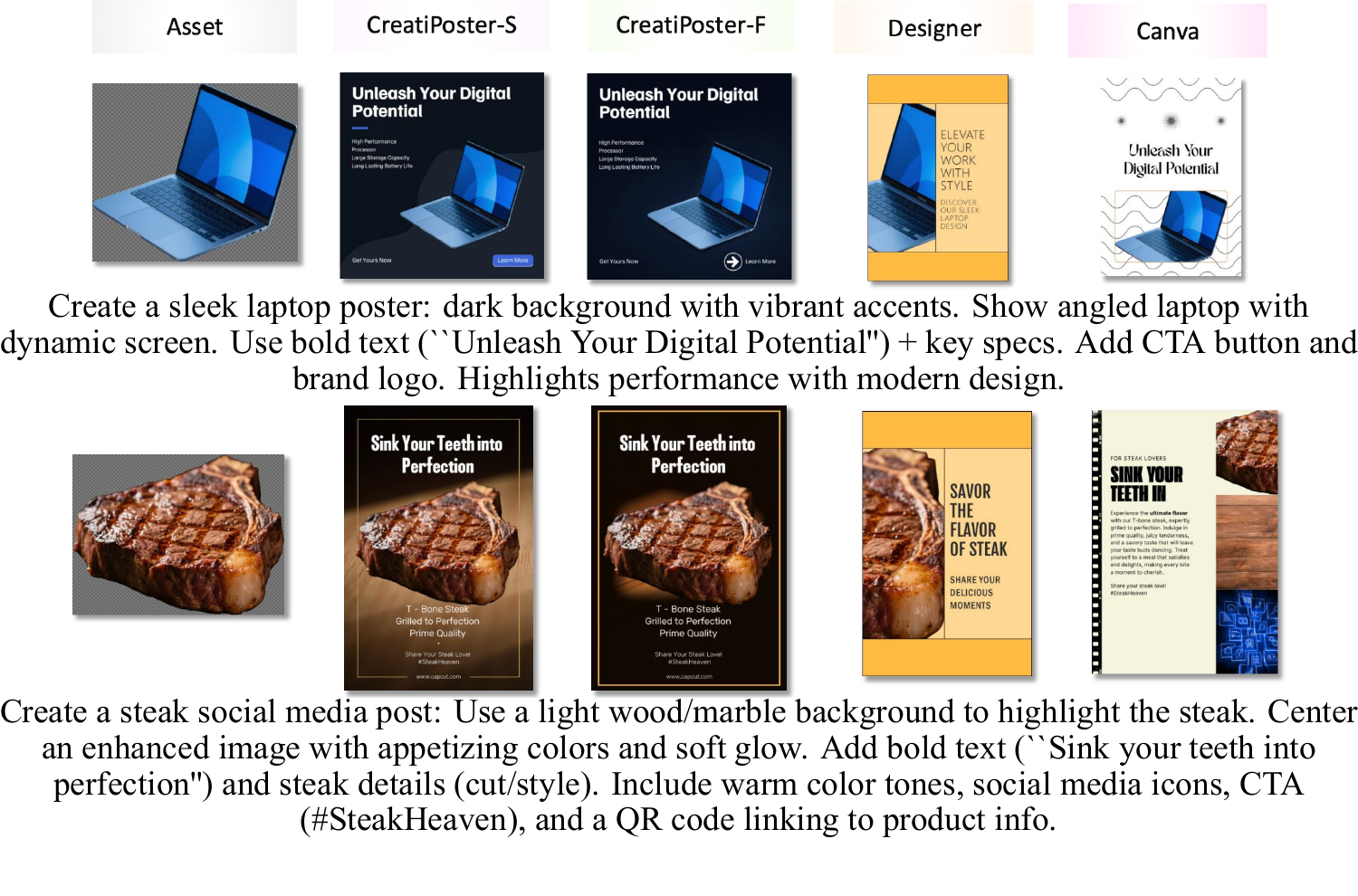}}
\end{subcaptionbox}
\caption{Qualitative comparison with OpenCOLE~\cite{inoue2024opencole}, Microsoft Designer, and Canva Magic Design.}
\label{fig:comp}
\end{figure}

We split the evaluation into a no-asset setting (\Cref{tab:comparison}) and an asset-conditioned setting (\Cref{tab:comparison_with_assets}); because the baselines do not support multiple-asset inputs, only the $84$ remaining cases enter the asset-conditioned comparison.
Representative qualitative comparisons against the baselines are shown in \Cref{fig:comp}.

\begin{table}[tbp]
\caption{Scores for Layout, Color, Graphic Style, and Compliance \emph{without} asset input.}
\label{tab:comparison}
\centering\small
\begin{tabular}{lccccc}
    \toprule
    \multicolumn{6}{c}{Scores from GPT-4.1} \\
    \midrule
    Method & \method-S & \method-F & OpenCOLE & MS Designer & Canva \\
    \midrule
    Layout & \textbf{2.89} & 2.71 & 1.60 & 2.61 & \underline{2.85} \\
    Color & 4.33 & \underline{4.36} & \textbf{4.57} & 3.55 & 4.11 \\
    Graphic Style & \textbf{4.24} & \underline{3.97} & 2.33 & 3.64 & 3.68 \\
    Compliance & \textbf{3.73} & \underline{3.67} & 3.03 & 2.38 & 3.20 \\
    \midrule
    \multicolumn{6}{c}{Scores from Human} \\
    \midrule
    Overall         & \textbf{2.80} & \underline{2.59} & 1.68 & 1.77 & 2.33 \\
    \bottomrule
\end{tabular}
\end{table}

\begin{table}[tbp]
\caption{Scores for Layout, Color, Graphic Style, and Compliance \emph{with} asset input.}
\label{tab:comparison_with_assets}
\centering\small
\begin{tabular}{lcccc}
    \toprule
    \multicolumn{5}{c}{Scores from GPT-4.1} \\
    \midrule
    Method & \method-S & \method-F & MS Designer & Canva \\
    \midrule
    Layout          & 2.41 & 2.25 & \underline{2.54} & \textbf{2.84} \\
    Color           & \textbf{4.28} & \underline{4.25} & \underline{4.25} & 4.02 \\
    Graphic Style   & \textbf{3.92} & \textbf{3.92} & 3.23 & 3.76 \\
    Compliance      & \textbf{3.59} & \underline{3.48} & 2.54 & 2.79 \\
    \midrule
    \multicolumn{5}{c}{Scores from Human} \\
    \midrule
    Overall         & \textbf{2.82} & \underline{2.67} & 2.05 & 2.33 \\
    \bottomrule
\end{tabular}
\end{table}

Across both tables \method-S/F take the top or near-top score on almost every dimension. \emph{Layout} remains the hardest dimension for all systems---no method exceeds $3.0$, reflecting a persistent gap versus human designers on complex, text-interwoven compositions. \emph{Color} scores are uniformly high, with GPT-4.1 favoring the harmonious palettes of OpenCOLE~\cite{inoue2024opencole} and \method-S. In \emph{Graphic Style}, \method{} leads overall, reflecting the value of jointly modeling text and asset typography via the layered protocol, and for \emph{Compliance} \method-S/F respond faithfully to prompts with and without assets. Human ratings agree with these trends; evaluators also noted that Canva Magic Design and Microsoft Designer often reuse template skeletons, yielding repetitive outputs that the layered protocol avoids. \method{} still exhibits two recurring failure modes---small-icon distortion and occasional text--asset misalignment---which we analyze in the supplementary material.

%% file: sections/application.tex
\section{Applications}
\label{sec:app}
\begin{figure}[!t]
    \centering
    \includegraphics[width=\textwidth]{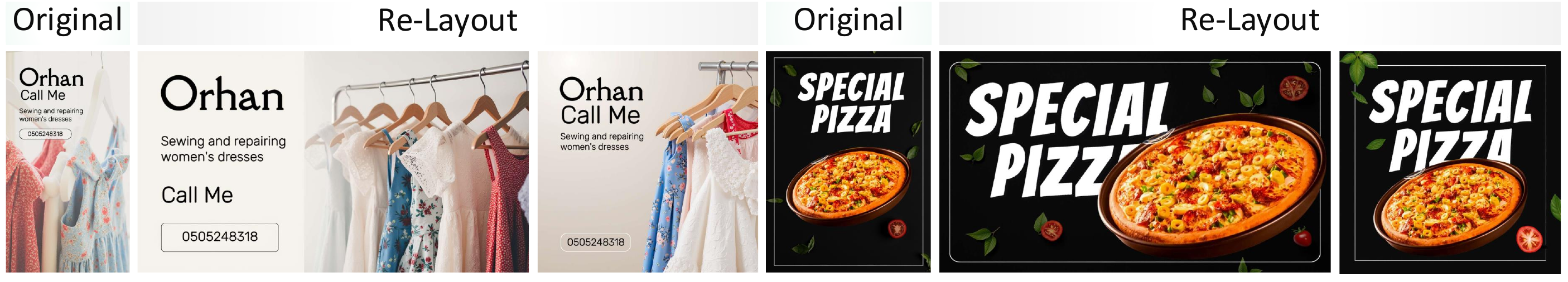}
    \caption{Poster re-layout. Given an original design, \method{} generates alternative layouts at various sizes while preserving content and style.}
    \label{fig:relayout}
\end{figure}

\begin{figure}[!t]
    \centering
    \includegraphics[width=\textwidth]{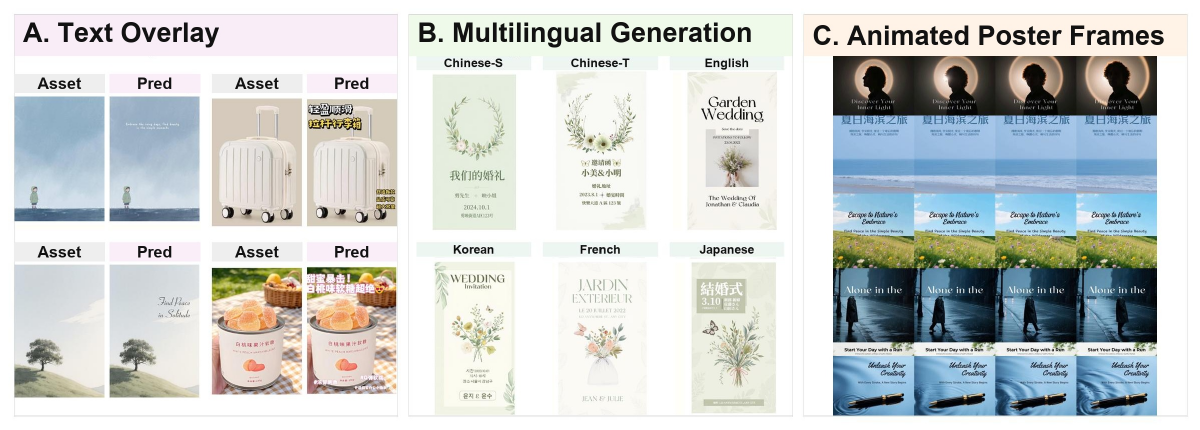}
    \caption{Additional applications of \method{}: (A) text overlay on uploaded assets, (B) multilingual generation, and (C) animated poster frames.}
    \label{fig:applications}
\end{figure}

Because \method{} outputs an editable design program rather than a flattened image, the single protocol-completion formulation supports diverse applications without task-specific changes (\Cref{fig:relayout,fig:applications}).
\textbf{Text overlay:} text is placed directly onto uploaded assets without the background synthesizer, e.g.\ titling product photos for e-commerce and social media.
\textbf{Poster re-layout:} reusing the rendered layers at a new canvas size, the generator predicts a new foreground and regenerates the background, yielding platform-specific posters or video covers from one design.
\textbf{Canvas mode:} canvas-conditioned completion lets users freeze chosen elements and fill in the rest, supporting multi-round editing; unlike closed-source Recraft\footnote{\url{https://www.recraft.ai/}}, our outputs stay layered and editable with better IP retention and text accuracy.
\textbf{Multilingual generation:} despite no multilingual protocol training, cross-lingual pre-training generalizes to Japanese, French, and Arabic.
\textbf{Animated poster:} since results are layered, text-to-video models~\cite{magi1} animate the background and merge it with the protocol-driven foreground, preserving text accuracy and editability.

%% file: sections/conclusion.tex
\section{Conclusion}
\label{sec:conclusion}
We presented \method, an open framework that shifts poster generation from pixel synthesis to \emph{editable design program synthesis}. A layered protocol generator trained under a single \emph{protocol completion} objective unifies prompt-only, asset-conditioned, canvas-editing, and re-layout generation, while a \emph{foreground-aware background synthesizer} composes coherent context around the rendered foreground without corrupting text or assets. \method{} outperforms leading open- and closed-source systems on automated and human evaluations, and we release code, model, a $100{,}000$-sample dataset, and a benchmark to advance accuracy, fidelity, editability, and aesthetics in one extensible stack.